\documentclass[10pt,twocolumn,letterpaper]{article}
\usepackage{cvpr}              

\usepackage{times}
\usepackage{epsfig}
\usepackage{graphicx}
\usepackage{amsmath}
\usepackage{amssymb}

\usepackage{arydshln}
\usepackage[dvipsnames,table,xcdraw]{xcolor}
\usepackage{soul}
\usepackage{url}
\usepackage{booktabs}
\usepackage{multirow}
\usepackage{float}
\usepackage{dblfloatfix}
\usepackage{wrapfig}
\usepackage{algorithm}
\usepackage{capt-of}
\usepackage[noend]{algpseudocode}
\usepackage[pagebackref,breaklinks,colorlinks]{hyperref}
\usepackage[capitalize]{cleveref}
\crefname{section}{Sec.}{Secs.}
\Crefname{section}{Section}{Sections}
\Crefname{table}{Table}{Tables}
\crefname{table}{Tab.}{Tabs.}

\newcommand{\ignore}[1]{}
\newcommand\T{\rule{0pt}{2.6ex}}       
\newcommand\sbullet[1][.5]{\mathbin{\vcenter{\hbox{\scalebox{#1}{$\bullet$}}}}}

\def\cvprPaperID{22} 
\def\confName{CVPR}
\def\confYear{2022}

\begin{document}

\title{Simple and Efficient Architectures for Semantic Segmentation}


\author{\vspace{-0.5cm}Dushyant Mehta \: Andrii Skliar \: Haitam Ben Yahia \: Shubhankar Borse 
\and 
\: Fatih Porikli \: Amirhossein Habibian \: Tijmen Blankevoort 
\\ Qualcomm AI Research\thanks{~Qualcomm AI Research is an initiative of Qualcomm Technologies,
Inc.}
\\ 
{\tt\small \{dushmeht,askliar,hyahia,sborse,fporikli,ahabibia,tijmen\}@qti.qualcomm.com}
\vspace{-0.5cm}
}
\maketitle
\begin{abstract}

Though the state-of-the architectures for semantic segmentation, such as HRNet, demonstrate impressive accuracy, the complexity arising from their salient design choices hinders a range of model acceleration tools, and further they make use of operations that are inefficient on current hardware. 
This paper demonstrates that a simple encoder-decoder architecture with a ResNet-like backbone and a small multi-scale head, performs on-par or better than complex semantic segmentation architectures such as HRNet, FANet and DDRNets.
 
 Na\"ively applying deep backbones designed for Image Classification to the task of Semantic Segmentation leads to sub-par results, owing to a much smaller effective receptive field of these backbones.
 Implicit among the various design choices put forth in works like HRNet, DDRNet, and FANet are networks with a large effective receptive field. 
 It is natural to ask if a simple encoder-decoder architecture would compare favorably if comprised of backbones that have a larger effective receptive field, though without the use of inefficient operations like dilated convolutions.

 We show that with minor and inexpensive modifications to ResNets, enlarging the receptive field, very simple and competitive baselines can be created for Semantic Segmentation.
 We present a family of such simple architectures for desktop as well as mobile targets, which match or exceed the performance of complex models on the Cityscapes dataset.  
 We hope that our work provides simple yet effective baselines for practitioners to develop efficient semantic segmentation models.  
 The model definitions and pre-trained weights are available at \url{https://github.com/Qualcomm-AI-research/FFNet}. 
\end{abstract}
\section{Introduction}
\begin{figure}[t!] 
\centering
  \includegraphics[width=1.0\linewidth]{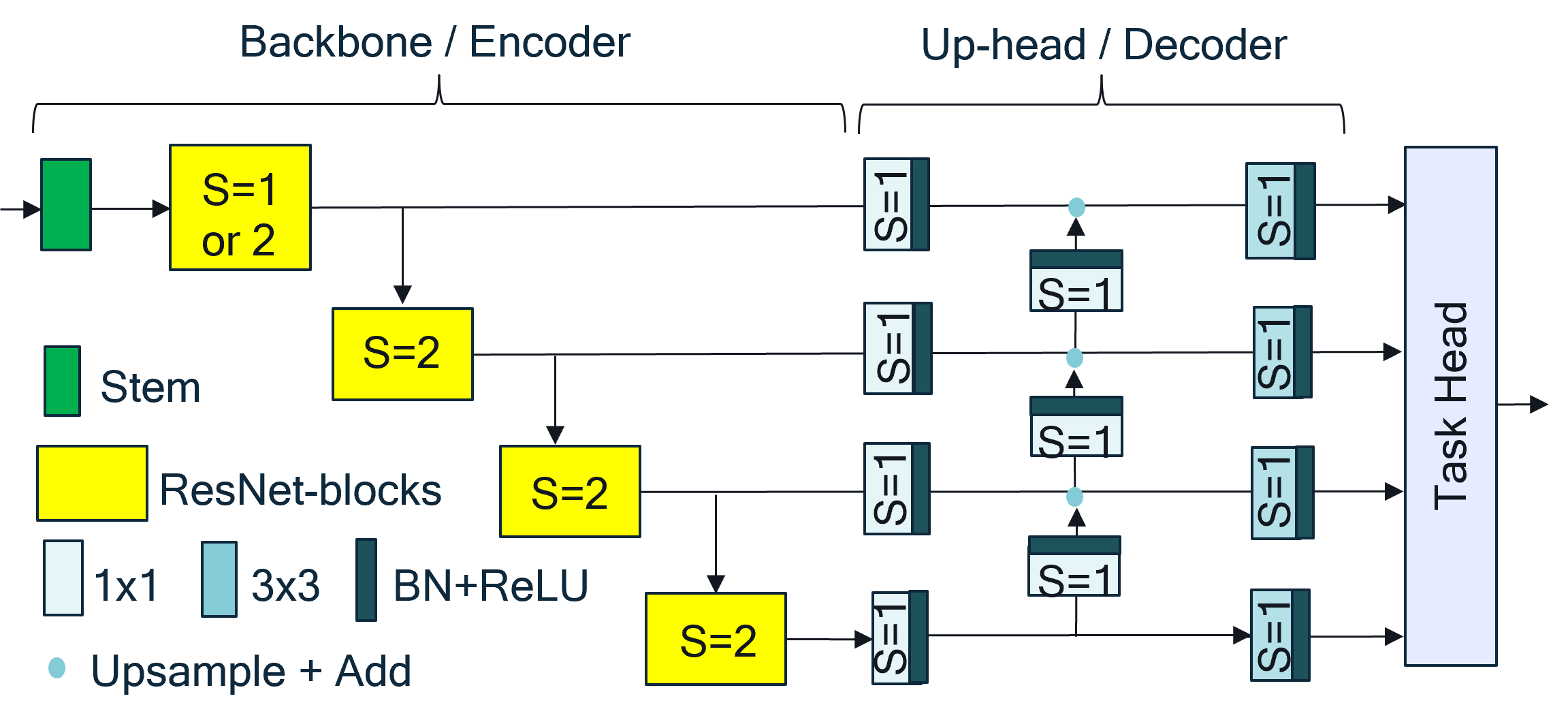}
  \caption{FFNet Architecture comprises of a backbone (encoder), in this case ResNet-like, feeding into a compact multi-branch Up-head (decoder), that subsequently feeds multi-scale features to the task specific head. `s' indicates the block or layer stride. The choices for the stem, widths \& depths of the backbone blocks, widths of the convolutions in the Up-head, choice of upsampling operator (bilinear vs nearest), and the design of the task head depend on the target platform and the task. The backbone options considered in this paper are listed in Table~\ref{tbl:ffnet_arch}, and the stem, Up-head and Segmentation-head options are depicted in Figure~\ref{fig:ffnet_options}.} 
\label{fig:ffnet_arch}
\end{figure}
\begin{figure*}[t!]
\centering
  \includegraphics[width=0.95\linewidth]{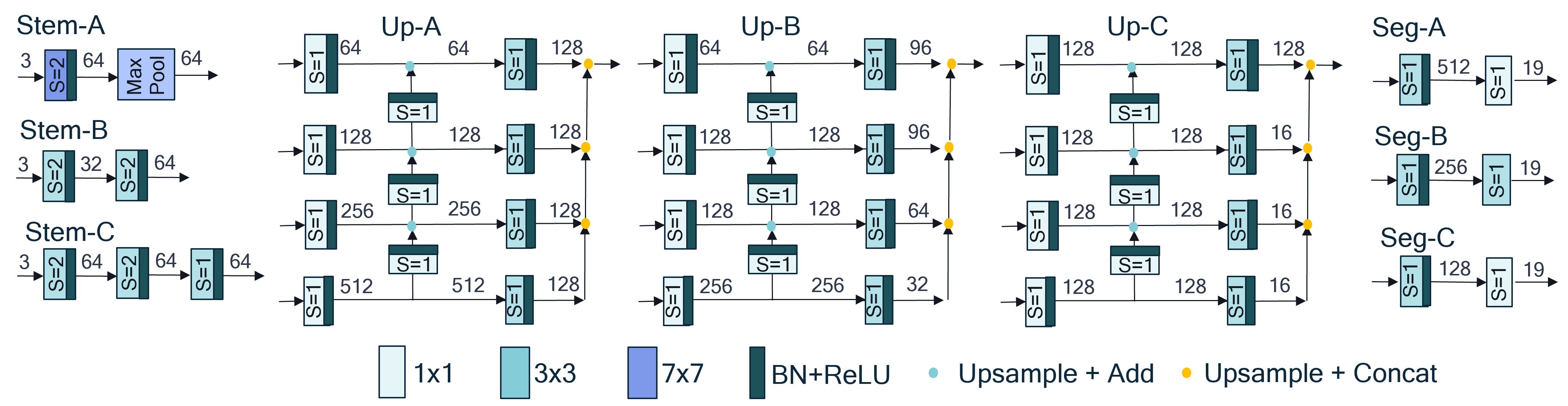}
  \caption{Various choices for the Stem, Up-head, and Segmentation-head that we consider in this paper. These are connected to backbone networks of varying widths and depths, listed in Table~\ref{tbl:ffnet_arch}. For GPU models, bilinear upsampling is used. For mobile models, nearest neighbour upsampling is used. The choices depicted are not comprehensive, and only meant to represent a few distinct points in a potential NAS search space. In the text we will refer to the Stem, Up-head and Segmentation-head choice combinations as A-B-B, C-B-C etc.}
\label{fig:ffnet_options}
\end{figure*}

Convolutional Neural Network based approaches to Semantic Image Segmentation have seen a lot of progress in the past years in terms of architectures, training techniques, datasets and loss functions. 
The extent of improvement arising from various aspects of the improved CNN architectures can often get muddled due to the specifics of training, datasets and training pipeline differences.

For semantic image segmentation of natural images, researchers and practitioners who are new to the field look to state-of-the-art tables~\cite{paperswithcode} and available code bases~\cite{nvidiasegmentation} as a starting point, and build on top~\cite{ocrnet19} of these.
Unfortunately, these benchmarks happen to be dominated by increasingly complex and costly architectures~\cite{hrnet,liteHRNet21,ddrnet21}.

We set out to examine the extent of efficacy of these complex designs over a conceptually simpler architecture. The baseline architecture we use is most closely related to FPN~\cite{FPN2017}, originally proposed for object detection. Specifically, we investigate whether the accuracy gains of the more complex architectures hold up when their associated non-architectural improvements are applied to this simple baseline architecture. We see that when using ResNet-50/101~\cite{resnet2016} backbones comprised of bottleneck-blocks in an FPN-like design the network is indeed markedly worse than the more complex designs. 
In this paper, we conjecture that this drop in performance is mainly due to the decrease in receptive field due to using bottleneck-blocks \cite{deeplab17}. We show that with similarly deep ResNet backbones comprised of basic-blocks, with a consequently much larger receptive field, the simple architecture does compare favorably to the more complex designs. Our simple architecture named \textit{FFNet} not only reduces the inference time and compute costs, 
but is comprised entirely of operations well supported on a wide variety of hardware, further making on-device deployment easier.

We describe the FFNet meta-architecture in Section~\ref{sec:ffnet}, and present results for a family of FFNet models in Section~\ref{sec:results}. 
Given the extent of the experiments and the compute requirements, we limit the experiments to the Cityscapes dataset.

\section{FFNet: A Simple Architecture}
\label{sec:ffnet}
In Figure~\ref{fig:ffnet_arch}, we depict a schema of our Fuss-Free Network (FFNet); a simple architecture inspired by the FPN architecture. It has an encoder-decoder structure. The encoder is comprised of a ResNet backbone without the classification head. Instead of just using features from the very last layer of the backbone, features at different spatial resolutions are extracted from all intermediate residual stages. These features are passed on to a light convolutional decoder head that upsamples and incorporates features from lower resolution branches to higher resolution branches. This decoder head, which we will refer to as `Up-head', also outputs features at different spatial resolutions. These multi-scale features are subsequently used as input to a small task-specific head, such as for segmentation or classification.

A lot of flexibility exists within the general setup of FFNet. We can freely change the width, depth, and the type of backbone building blocks, number of feature scales, the head type and the head width.
Figure~\ref{fig:ffnet_options} depicts the various stem, Up-head, and Segmentation-head choices we investigate in this paper, marked as A/B/C. These are combined with various backbone width and depth configurations depicted in Table~\ref{tbl:ffnet_arch}, depending on the target platform.
The first residual block in the backbone can have a stride of 1 or 2, which changes the spatial resolution of the output. We show results for both in Section~\ref{sec:results}. 
Specific instantiations of FFNets throughout the paper will refer to a backbone, such as ResNet-50, paired with a Stem-Up-Seg combination denoted as A-B-B.

We show models for desktop GPU and mobile targets. The desktop models use bilinear upsampling in the Up-head, while mobile models use nearest neighbour upsampling.  


This is a simple design with no particular restrictions imposed on the number of stages in the backbone or the number of feature scales being output, the choice of striding in the stem and the backbone. The architecture can thus easily be adapted for other tasks. 
%
\begin{table*}[]
\setlength{\tabcolsep}{3pt}
\caption{Configurations of backbones examined in the paper. The naming convention follows that of ResNets~\cite{resnet2016}, even though these are paired up with different stems and heads depicted in Figure~\ref{fig:ffnet_options}, which changes the overall number of layers.}
\label{tbl:ffnet_arch}
\begin{tabular}{l|l|c|c|c|c}
\multicolumn{1}{c|}{} & \multicolumn{1}{c|}{Network Name} & ResNet Blocks & Stage-wise Num Blocks & Stage Output Channels & Stage Strides         \\ \hline
                                 4-stage Backbones & ResNet 150          & Basic         & {[}16, 18, 28, 12{]}   & {[}64, 128, 256, 512{]}    & {[}1 or 2, 2, 2, 2{]} \\
                                  & ResNet 134          & Basic         & {[}8, 18, 28, 12{]}    & {[}64, 128, 256, 512{]}    & {[}1 or 2, 2, 2, 2{]} \\
                                  & ResNet 101          & Bottleneck    & {[}3, 4, 23, 3{]}      & {[}256, 512, 1024, 2048{]} & {[}1 or 2, 2, 2, 2{]} \\
                                  & ResNet 86           & Basic         & {[}8, 12, 16, 6{]}     & {[}64, 128, 256, 512{]}    & {[}1 or 2, 2, 2, 2{]} \\
                                  & ResNet 56           & Basic         & {[}4, 8, 12, 3{]}      & {[}64, 128, 256, 512{]}    & {[}1 or 2, 2, 2, 2{]} \\
                                  & ResNet 50           & Bottleneck    & {[}3, 4, 6, 3{]}       & {[}256, 512, 1024, 2048{]} & {[}1 or 2, 2, 2, 2{]} \\
                                  & ResNet 34           & Basic         & {[}3, 4, 6, 3{]}       & {[}64, 128, 256, 512{]}    & {[}1 or 2, 2, 2, 2{]} \\
                                  & ResNet 18           & Basic         & {[}2, 2, 2, 2{]}       & {[}64, 128, 256, 512{]}    & {[}1 or 2, 2, 2, 2{]} \\ \cline{2-6}
                                  & ResNet 150 S         & Basic         & {[}16, 18, 28, 12{]}   & {[}64, 128, 192, 320{]}    & {[}1 or 2, 2, 2, 2{]} \\
                                  & ResNet 86 S          & Basic         & {[}8, 12, 16, 6{]}     & {[}64, 128, 192, 320{]}    & {[}1 or 2, 2, 2, 2{]} \\
                                  & ResNet 78 S          & Basic         & {[}6, 12, 12, 8{]}     & {[}64, 128, 192, 320{]}    & {[}1 or 2, 2, 2, 2{]} \\
                                  & ResNet 54 S          & Basic         & {[}5, 8, 8, 5{]}       & {[}64, 128, 192, 320{]}    & {[}1 or 2, 2, 2, 2{]} \\
                                  & ResNet 40 S          & Basic         & {[}4, 5, 6, 4{]}       & {[}64, 128, 192, 320{]}    & {[}1 or 2, 2, 2, 2{]} \\
                                  & ResNet 30 S          & Basic         & {[}3, 4, 4, 3{]}       & {[}64, 128, 192, 320{]}    & {[}1 or 2, 2, 2, 2{]} \\
                                  & ResNet 22 S          & Basic         & {[}2, 3, 3, 2{]}       & {[}64, 128, 192, 320{]}    & {[}1 or 2, 2, 2, 2{]} \\ \hline
                                 3-Stage Backbones & ResNet 122 N          & Basic         & {[}16, 24, 20{]}        & {[}96, 160, 320{]}         & {[}2, 2, 2{]}        \\
                                  & ResNet 74 N          & Basic         & {[}8, 12, 16{]}        & {[}96, 160, 320{]}         & {[}2, 2, 2{]}   \\ 
                                  & ResNet 46 N          & Basic         & {[}6, 8, 8{]}          & {[}96, 160, 320{]}         & {[}2, 2, 2{]}         \\ \cline{2-6}
                                  & ResNet 122 NS           & Basic         & {[}16, 24, 20{]}        & {[}64, 128, 256{]}         & {[}2, 2, 2{]}         \\
                                  & ResNet 74 NS           & Basic         & {[}8, 12, 16{]}        & {[}64, 128, 256{]}         & {[}2, 2, 2{]}         \\
                                  & ResNet 46 NS          & Basic         & {[}6, 8, 8{]}          & {[}64, 128, 256{]}         & {[}2, 2, 2{]}         \\ \hline
\end{tabular}
\end{table*}

\subsection{Experimental Setup}
\textbf{Dataset.} The Cityscapes~\cite{cordts2016cityscapes} dataset is comprised of videos of urban street scenes recorded from a moving vehicle. 5k image frames have high quality pixel-level annotations of 19 semantic classes, and a further 20k frames that are weakly annotated.

\textbf{Training Setup.}
Since the number of annotated frames in the Cityscapes dataset is limited, we utilize ImageNet pretraining of the models. The FFNet models are pretrained on the ILSVRC2012~\cite{deng2009imagenet} dataset in the FFNet form, using the classification head from HRNet~\cite{hrnet} in place of the segmentation head. 
For ImageNet pretraining, we make use of Pytorch Image Models~\cite{rw2019timm}, and train for 150 epochs.

For Cityscapes training, the stem and the backbone are initialized from the ImageNet pretraining. The Up-head and the Segmentation-head are initialized from scratch. 
Since the fine grained annotations on the Cityscapes dataset are limited to ~5k images, learning meaningful and generalizable feature representations with deep CNNs can be challenging. We use pseudo labels from Tao et al.~\cite{hierarchicalMultiScale2020} on the images with coarse annotations. %
Additionally, we use the RMI~\cite{rmiLoss2019} loss as an auxiliary loss while training. The models are trained for 175 epochs with a batch size of 8. 
We will make the ImageNet and Cityscapes training hyperparameters available with the models.



\section{Experiments and Comparisons}
\label{sec:results}
We examine various configurations of FFNets. The backbones of different depths, widths, and block types that we tried are detailed in Table~\ref{tbl:ffnet_arch}. These backbones are paired with various stem, Up-head, and Segmentation options from Figure~\ref{fig:ffnet_options}, and the configurations referred to as A-B-B etc.

For desktop GPU FFNets, we use input images of size $1024\times2048$. We show FFNet models with a stride of 1 and 2 in the first residual stage of the ResNet backbone, with an output segmentation map resolution of $256\times512$ and $128\times256$ respectively. We'll refer to these models as FFNet-GPU-Large and FFNet-GPU-Small respectively.

For mobile FFNets, we use a stride of 2 in the first residual stage, with an input resolution of $1024\times2048$, and a stride of 1 when the input resolution is $512\times1024$, both resulting in a segmentation map resolution of $128\times256$.
\begin{table}[]
\setlength{\tabcolsep}{1pt}
\caption{
Models with an output segmentation map resolution of $256\times512$ for an input image resolution of $1024\times2048$. The FFNets have stride=1 in the first block of the ResNet backbone. The timings reported here are without batch norm folding, and a batch size of 1. See the associated plots in Figure~\ref{fig:stride_32_models}}
\label{tbl:gpu_large}
\begin{tabular}{cc|ccc|c}
\multicolumn{2}{c|}{\multirow{3}{*}{Model}}  & \multicolumn{3}{c|}{Inference Time (ms)}                                   & Citysc.           \\
\multicolumn{2}{c|}{}                        & \multicolumn{1}{c|}{1080Ti} & \multicolumn{2}{c|}{2080Ti}                  & Val                  \\  
\multicolumn{2}{c|}{}                        & \multicolumn{1}{c|}{FP32}   & FP32                 & FP16                  & mIoU                 \\ \hline
\multicolumn{2}{c|}{\textcolor{red}{$\blacktriangle$}HRNetv2 48}             & \multicolumn{1}{c|}{249}    & 171                  & 122                   & 84.9                 \\
\multicolumn{2}{c|}{\textcolor{red}{$\blacktriangle$}HRNetv2 32}             & \multicolumn{1}{c|}{140}    & 100                  & 77                    & 83.7                 \\
\multicolumn{2}{c|}{\textcolor{red}{$\blacktriangle$}HRNetv2 18}             & \multicolumn{1}{c|}{97}     & 69                   & 61                    & 82.4                 \\ \hline
\multicolumn{2}{c|}{\T\textbf{FFNet-GPU-Large}}                  & \multicolumn{1}{l|}{}       & \multicolumn{1}{l}{} & \multicolumn{1}{l|}{} & \multicolumn{1}{l}{} \\ 
Backbone             & Stem-Up-Seg           & \multicolumn{1}{c|}{}       &                      &                       &                      \\ \hline
\multicolumn{1}{l}{\textcolor{blue}{$\sbullet[2.0]$}ResNet 150}           & A-A-A                 & \multicolumn{1}{c|}{195}    & 152                  & 81                    & 84.4                 \\
\multicolumn{1}{l}{\textcolor{blue}{$\sbullet[2.0]$}ResNet 134}           & A-A-A                 & \multicolumn{1}{c|}{176}    & 135                  & 70                    & 84.1                 \\
\multicolumn{1}{l}{\textcolor{blue}{$\sbullet[2.0]$}ResNet 86}            & A-A-A                 & \multicolumn{1}{c|}{139}    & 105                  & 55                    & 83.2                 \\
\multicolumn{1}{l}{\textcolor{blue}{$\sbullet[2.0]$}ResNet 56}            & A-A-A                 & \multicolumn{1}{c|}{112}    & 82                   & 42                    & 82.5                 \\
\multicolumn{1}{l}{\textcolor{blue}{$\sbullet[2.0]$}ResNet 34}            & A-A-A                 & \multicolumn{1}{c|}{95}     & 67                   & 34                    & 81.4                 \\ \hline
\multicolumn{1}{l}{\textcolor{violet}{$\sbullet[2.0]$}ResNet 101}           & A-A-A                 & \multicolumn{1}{c|}{181}    & 119                  & 59                    & 82.1                 \\
\multicolumn{1}{l}{\textcolor{violet}{$\sbullet[2.0]$}ResNet 50}            & A-A-A                 & \multicolumn{1}{c|}{134}    & 88                   & 45                    & 79.6                 \\ \hline
\multicolumn{1}{l}{\textcolor{cyan}{$\sbullet[2.0]$}ResNet 150}           & A-B-B                 & \multicolumn{1}{c|}{151}    & 125                  & 71                    & 83.8                 \\
\multicolumn{1}{l}{\textcolor{cyan}{$\sbullet[2.0]$}ResNet 86}            & A-B-B                 & \multicolumn{1}{c|}{97}     & 78                   & 45                    & 83.5                 \\
\multicolumn{1}{l}{\textcolor{cyan}{$\sbullet[2.0]$}ResNet 56}            & A-B-B                 & \multicolumn{1}{c|}{70}     & 56                   & 32                    & 82.1                 \\
\multicolumn{1}{l}{\textcolor{cyan}{$\sbullet[2.0]$}ResNet 34}            & A-B-B                 & \multicolumn{1}{c|}{52}     & 41                   & 25                    & 80.3                 \\ \hline
\multicolumn{1}{l}{\textcolor{BlueViolet}{$\sbullet[2.0]$}ResNet 150 S}     & B-B-B                 & \multicolumn{1}{c|}{126}    & 104                  & 66                    & 84.1                     \\
\multicolumn{1}{l}{\textcolor{BlueViolet}{$\sbullet[2.0]$}ResNet 86 S}     & B-B-B                 & \multicolumn{1}{c|}{83}     & 67                   & 43                    & 82.6                     \\ \hline
\multicolumn{1}{l}{\textcolor{Green}{$\blacksquare$}ResNet 122 N}          & C-B-B                 & \multicolumn{1}{c|}{71}     & 58                   & 44                    & 83.6                     \\
\multicolumn{1}{l}{\textcolor{Green}{$\blacksquare$}ResNet 74 N}          & C-B-B                 & \multicolumn{1}{c|}{52}     & 42                   & 32                    & 83.0   \\               
\multicolumn{1}{l}{\textcolor{Green}{$\blacksquare$}ResNet 46 N}          & C-B-B                 & \multicolumn{1}{c|}{43}     & 34                   & 27                    & 81.9                  
\end{tabular}
\end{table}

\begin{figure*}
\centering
  \includegraphics[width=1.0\linewidth]{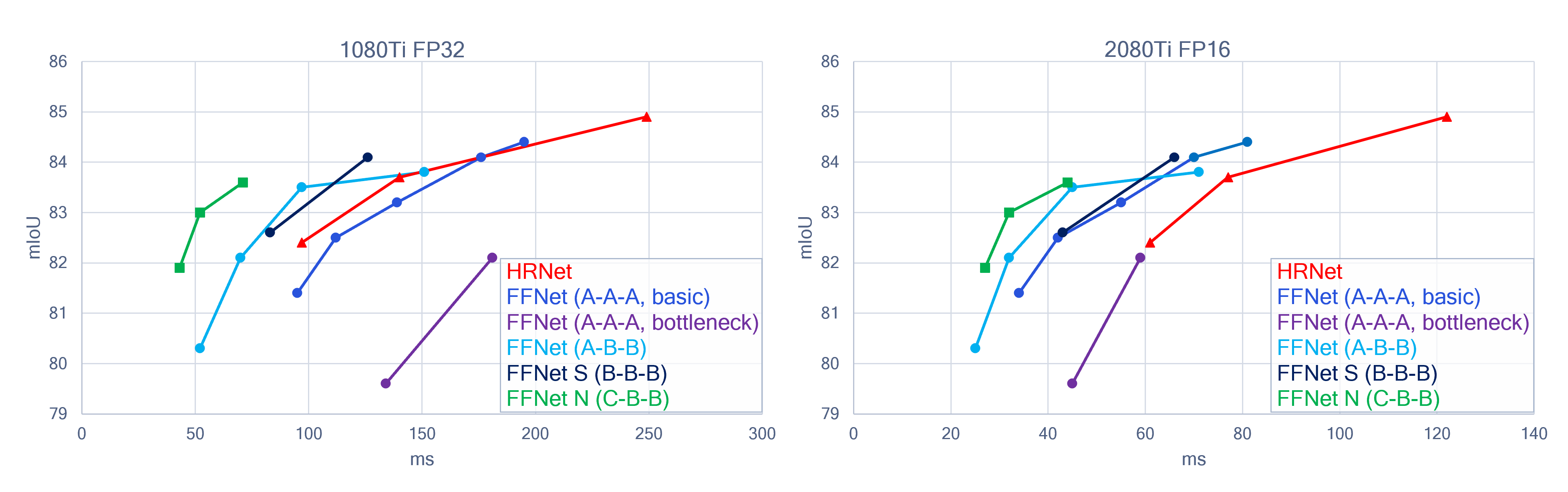}
  \caption{Inference latency vs Cityscapes validation accuracy for \textbf{FFNet-GPU-Large Models}: Simple FFNets (blue) using basic-blocks are on par with HRNets (red), while FFNets (violet) with bottleneck-blocks are markedly worse. Exploring various combinations of Up-head widths (cyan) and backbone widths (black), can allow better models to be created. FFNets with a 3-stage backbone (green) may offer better pareto performance than 4-stage FFNets. Input resolution $1024\times2048$, output segmentation map resolution $256\times512$. See Table~\ref{tbl:gpu_large}.}
\label{fig:stride_32_models}
\end{figure*}
\subsection{Efficacy of Basic Blocks vs Bottleneck Blocks}
FFNets with backbones comprised of basic-blocks far outperform backbones comprised of bottleneck blocks, for FFNet-GPU-Large and FFNet-GPU-Small models. See FFNet A-A-A in Figure~\ref{fig:stride_32_models} and Figure~\ref{fig:stride_64_models}. 
Prior work has primarily used ResNet-101 and ResNet-50 as backbones in baselines for Segmentation, and seen poor results on account of an inadequately large receptive field, and needing to enlarge the receptive field through tricks like dilated convolution~\cite{deeplab17}. New backbones like ResNet-150, 134, 86, 56 described in Table~\ref{tbl:ffnet_arch}, and known backbones like ResNet-34, which are all comprised of basic-blocks fare far better than ResNet-50 and ResNet-101, particularly in the case of FFNet-GPU-Large which operates on larger feature maps.

\subsection{Exploring the Model Space}
While the effects of varying the backbone depth are obvious and expected, we also experiment with varying the width of the backbone, and the width of the Up-head and the Segmentation-head. 
For FFNet-GPU-Large models, given the larger spatial resolution of the output feature maps, significant speed improvements arise from narrowing the Up-head and the Segmentation-head. FFNet A-B-B (cyan) in Figure~\ref{fig:stride_32_models}, which uses narrower Up-head and Segmentation-head, provides a better speed-accuracy trade-off than FFNet A-A-A (blue) with its wider heads. 
For FFNet-GPU-Small models, narrowing the Up-head and the Segmentation head offers minor latency improvements as seen with FFNet A-A-C (cyan) vs FFNet A-A-A (blue) in Figure~\ref{fig:stride_64_models} and FFNet A-B-B vs FFNet A-A-C in Table~\ref{tbl:gpu_small}.
Narrowing the backbone can help further on resource constrained devices. On desktop GPUs, however, there may not be a noticeable advantage. FFNet S B-B-B, comprised of narrower backbones, shows similar performance as FFNet A-B-B, that use a wider backbone, as shown in Figures~\ref{fig:stride_32_models},~\ref{fig:stride_64_models}.

\subsection{Models for Mobile}
\begin{figure}
\centering
  \includegraphics[trim={0 0 0 0cm},clip,width=1.0\linewidth]{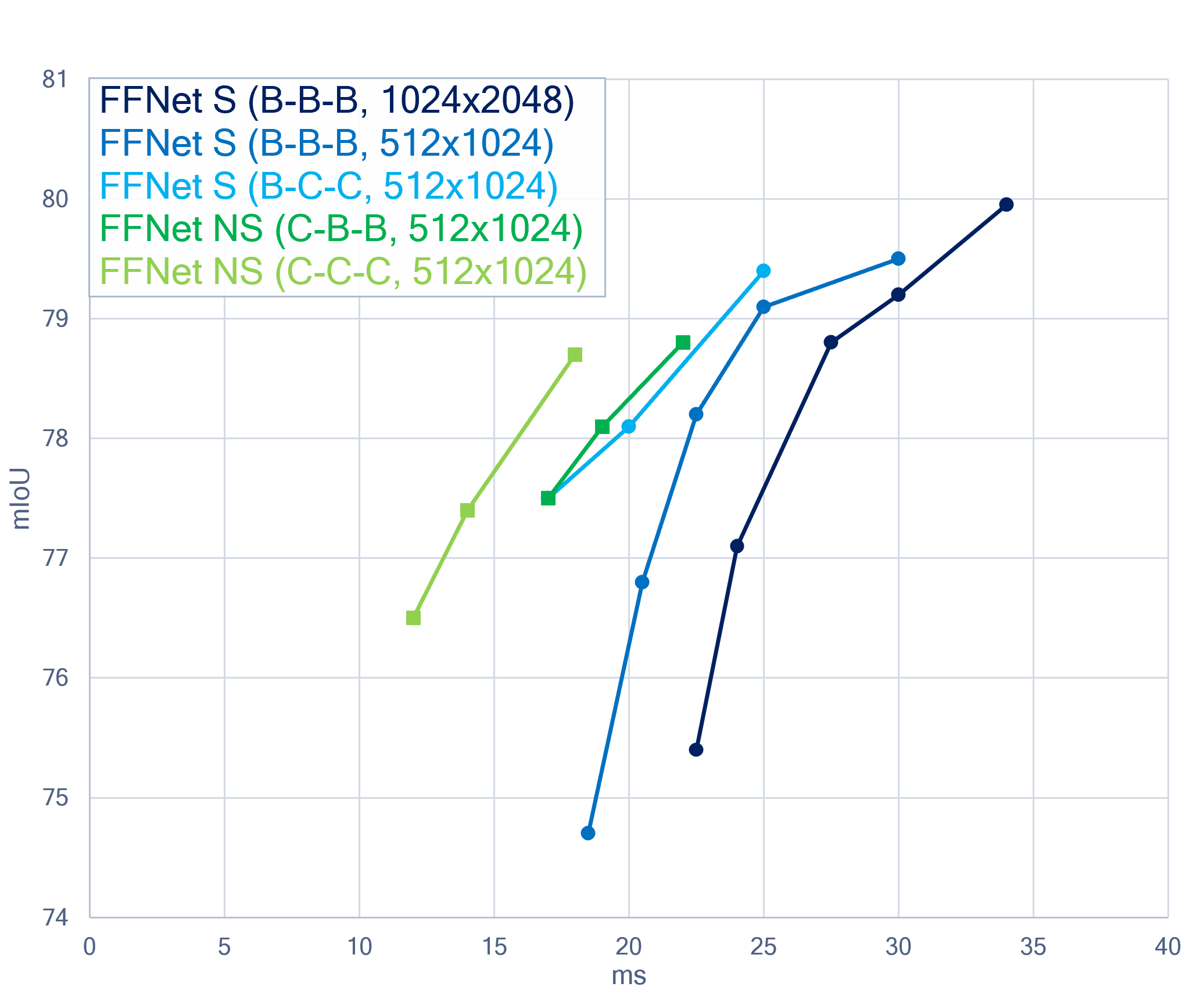}
  \caption{Inference latency vs Cityscapes validation accuracy for \textbf{FFNet-Mobile Models} on Samsung S21: The models run in real-time on the mobile DSP. Regardless of input resolution, all models output segmentation maps at the same resolution. Using a smaller input resolution (blue, cyan) affords a better speed-accuracy tradeoff than a higher input resolution with additional downsampling in the backbone (black). Narrowing the head B-B-B (blue) $\rightarrow$ B-C-C (cyan) offers a better speed-accuracy tradeoff. As for desktop GPUs, 3-stage backbones (green, light green) are more efficient than 4-stage models for mobile too. See Table~\ref{tbl:device}.}
\label{fig:s21_models}
\end{figure}

For FFNet models intended for mobile deployment, we utilize narrower ResNet backbones and heads. Instead of bilinear upsampling in the heads, FFNet-Mobile models use nearest neighbour upsampling.
We show a family of models with their inference times on a Samsung S21 DSP, in Figure~\ref{fig:s21_models} and Table~\ref{tbl:device}.
Changing the input resolution, and the width of the head allows on-device models to be created with a better speed-accuracy tradeoff. See black, blue and cyan plots in Figure~\ref{fig:s21_models}.

\subsection{3-Stage Backbone}
\begin{figure}
\centering
  \includegraphics[width=1.0\linewidth]{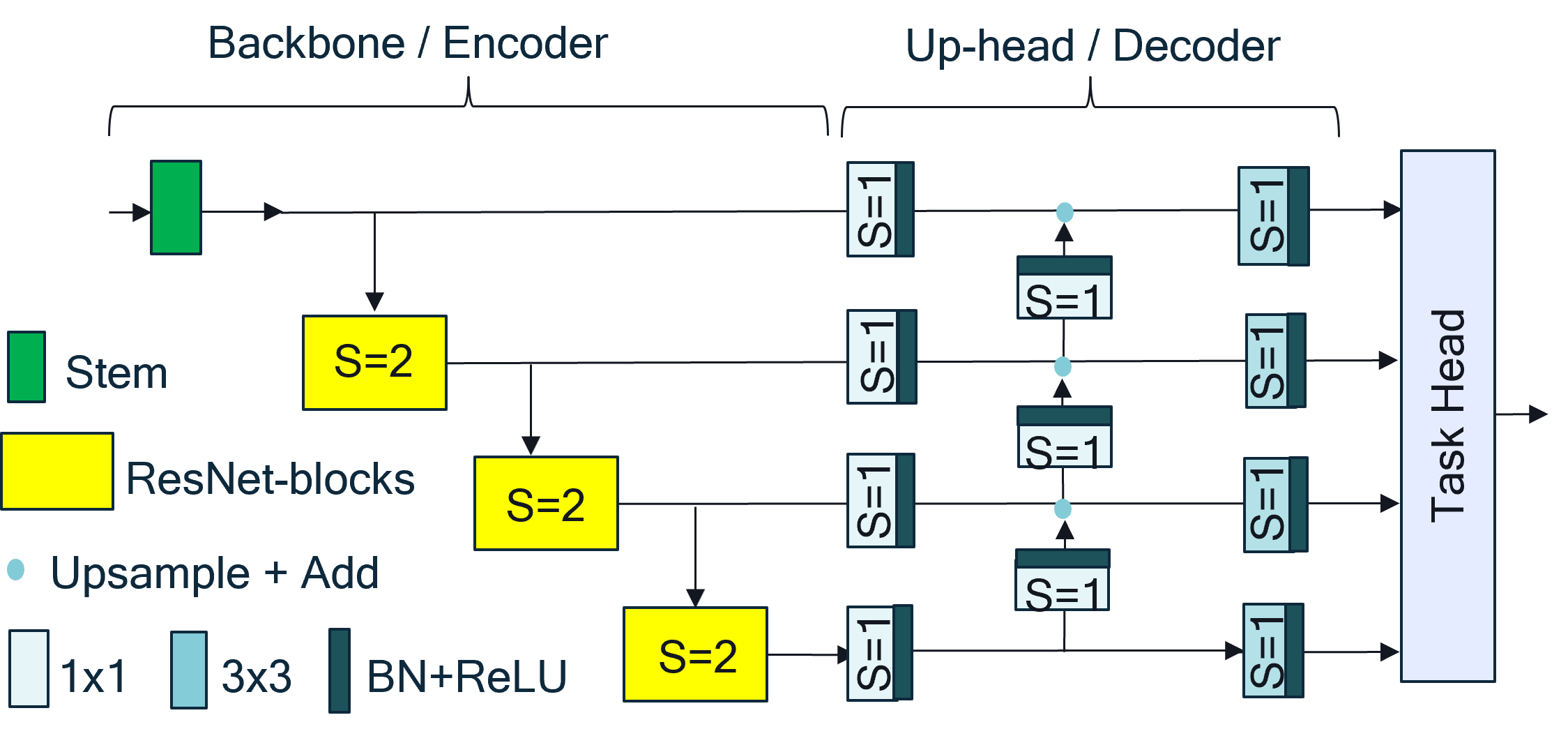}
  \caption{FFNet variant with a 3-stage backbone.}
\label{fig:ffnet_3_stage_arch}
\end{figure}

We experiment with the backbone structure, beyond the depth and width of the backbone, in order to create a more efficient FFNet. Typical ResNets are comprised of 4 residual stages. The first stage operates on the largest feature maps, and has a significant compute and memory cost. We create a 3-stage ResNet backbone, removing the first residual stage. For FFNets, the 3-stage backbone interfaces with the Up-head, using the output from the stem instead of the output from the first residual stage of a 4-stage backbone. See Figure~\ref{fig:ffnet_3_stage_arch}.
We see that across the tested devices, FFNets with a 3-Stage backbone, can offer a better speed-accuracy trade-off than FFNets with a 4-stage backbone for certain accuracy ranges. See FFNet N C-B-B (green) in Figures~\ref{fig:stride_32_models},~\ref{fig:stride_64_models} for GPU models, and FFNet NS C-B-B (green) and FFNet NS C-C-C (light green) in Figure~\ref{fig:s21_models}. 

It is unclear if the 3 stage model would also be better at higher accuracies, but this experimentation reveals a new dimension of exploration to be covered in a potential NAS search space.

\subsection{Comparisons to Prior Art}
\label{sec:prior_art}
We use HRNet~\cite{hrnet}, DDRNet~\cite{ddrnet21}, and FANet~\cite{fanet2020} as our points of comparison to prior work. HRNets are commonplace in much of semantic segmentation work. DDRNet demonstrates better performance than several classes of GPU efficient networks for Semantic Segmentation. FANets also claim GPU efficiency, and are the closest to our design, but make use of self-attention in the head.
We compare the accuracy and GPU inference timings for the models, and refrain from making FLOP comparisons for a few reasons. Various libraries for FLOP computation are not able to correctly handle all kinds of operations, and may need operations implemented a certain way to correctly account for the FLOPs. Further, several operations such as tensor concatenation and channel shuffling do not have a compute cost associated, but incur a large memory access time cost. Hence, we restrict ourselves to inference time comparisons.
We retrain all the models that we compare to, using the same training scheme,  to make comparisons fair. Hence, the performance reported for these models is much higher than that in their respective papers.
However, there may yet be better accuracy possible with model specific training-hyperparameter tuning. The validation accuracy is given for single scale inference, with no test-time augmentation used.

\textbf{Comparison with HRNet}: 
HRNet utilizes multi-scale feature representations throughout the network, with repeated information exchange across the feature scales, and a multi-scale feature output that is used by the task specific head. 
FFNet A-A-A (blue) models using backbones comprised of basic-blocks perform on par with HRNets (red), as shown in Figure~\ref{fig:stride_32_models}, while FFNet A-A-A (violet) models using bottleneck-blocks are markedly worse. 
A key feature of HRNet that is not remarked upon, is that the network is 75+ layers deep, and comprised of basic-blocks. It consequently has a larger effective receptive field than bottleneck-block based networks like ResNet-50/101. 
We find that a multi-scale feature output constructed with the simple FFNet  architectures achieves similar accuracy, when the backbones are made of basic-blocks.
These results point to the fact that the large receptive field is a critical contributor to the overall accuracy, and comparable results are possible with simple designs like the FFNets, without requiring other components such as a repeated information exchange across scales.
It would need to be determined when and if the repeated information exchange across feature scales is effective. 
Other families of FFNets with narrower heads and backbones can perform better than HRNets as seen in Table~\ref{tbl:gpu_large}.

\begin{figure*}
\centering
  \includegraphics[width=1.0\linewidth]{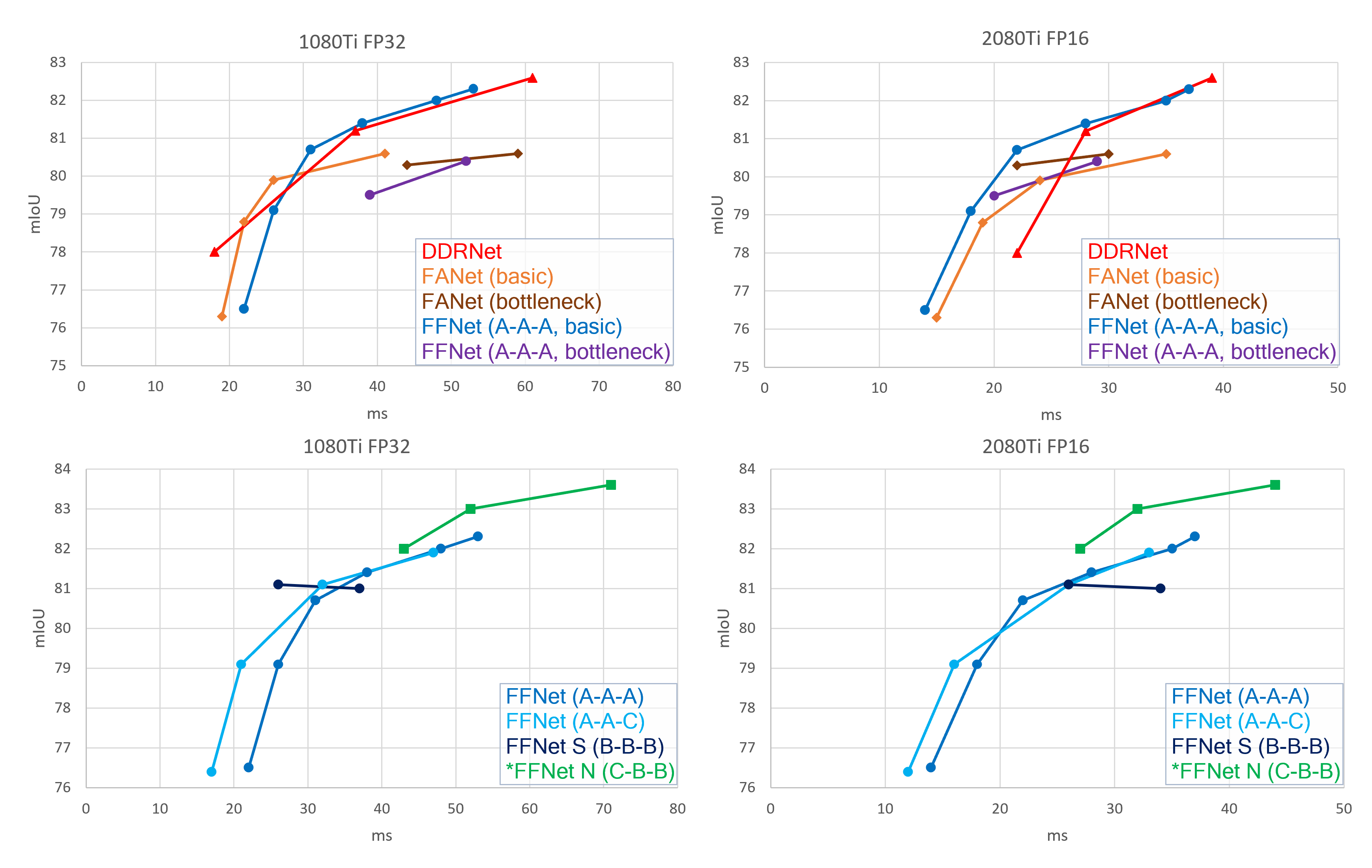}
  \caption{Inference latency vs Cityscapes validation accuracy for \textbf{FFNet-GPU-Small Models}: (Top) FFNets (blue) perform on par with complex models like DDRNet (red) and  FANet (orange and brown). For FFNets, basic-block (blue) is always better than bottleneck-block (violet). (Bottom) The model space can be explored via combinations of basic-block backbones and Up-heads of various widths (cyan and black), to create better models. 
  Input resolution $1024\times2048$, output segmentation map resolution $128\times256$. See Table~\ref{tbl:gpu_small}.
  Also shown here are 3-stage FFNet-GPU-Large (green) models, which far outperform 4-stage FFNet-GPU-Small models.
  }
\label{fig:stride_64_models}
\end{figure*}

\begin{table}[]
\setlength{\tabcolsep}{1pt}
\caption{
Models with an output segmentation map resolution of $128\times256$ for an input image resolution of $1024\times2048$. The timings reported here are without batch norm folding, and a batch size of 1. 
$\dagger$ These models use stride=2 in the first stage of the ResNet backbone.
See the associated plots in Figure~\ref{fig:stride_64_models}.
}
\label{tbl:gpu_small}
    \begin{tabular}{lc|ccc|c}
\multicolumn{2}{c|}{\multirow{3}{*}{Model}} & \multicolumn{3}{c|}{Inference Time (ms)}                  & Citysc. \\
\multicolumn{2}{c|}{}                       & \multicolumn{1}{c|}{1080Ti} & \multicolumn{2}{c|}{2080Ti} & Val        \\
\multicolumn{2}{c|}{}                       & \multicolumn{1}{c|}{FP32}   & FP32         & FP16         & mIoU       \\ \hline
\multicolumn{2}{l|}{\textcolor{red}{$\blacktriangle$}DDRNet 39}              & \multicolumn{1}{c|}{61}     & 47           & 39           & 82.6       \\
\multicolumn{2}{l|}{\textcolor{red}{$\blacktriangle$}DDRNet 23}              & \multicolumn{1}{c|}{37}     & 27           & 28           & 81.2       \\
\multicolumn{2}{l|}{\textcolor{red}{$\blacktriangle$}DDRNet 23 S}         & \multicolumn{1}{c|}{18}     & 17           & 22           & 78.0         \\ \hline
\multicolumn{2}{l|}{\textcolor{YellowOrange}{$\blacklozenge$}$\dagger$ FANet 134}              & \multicolumn{1}{c|}{41}     & 34           & 35           & 80.6       \\
\multicolumn{2}{l|}{\textcolor{YellowOrange}{$\blacklozenge$}$\dagger$ FANet 56}               & \multicolumn{1}{c|}{26}     & 23           & 24           & 79.9       \\
\multicolumn{2}{l|}{\textcolor{YellowOrange}{$\blacklozenge$}$\dagger$ FANet 34}               & \multicolumn{1}{c|}{22}     & 19           & 19           & 78.8       \\
\multicolumn{2}{l|}{\textcolor{YellowOrange}{$\blacklozenge$}$\dagger$ FANet 18}               & \multicolumn{1}{c|}{19}     & 17           & 15           & 76.3       \\ \cline{1-2}
\multicolumn{2}{l|}{\textcolor{Bittersweet}{$\blacklozenge$}$\dagger$ FANet 101}              & \multicolumn{1}{c|}{59}     & 36           & 30           & 80.6       \\
\multicolumn{2}{l|}{\textcolor{Bittersweet}{$\blacklozenge$}$\dagger$ FANet 50}               & \multicolumn{1}{c|}{44}     & 28           & 22           & 80.3       \\ \hline 
\multicolumn{2}{c|}{\T\textbf{FFNet-GPU-Small}}                 & \multicolumn{1}{c|}{}       &              &              &            \\ 
Backbone               & Stem-Up-Seg        & \multicolumn{1}{c|}{}       &              &              &            \\ \hline
\textcolor{blue}{$\sbullet[2.0]$}$\dagger$ ResNet 150             & A-A-A              & \multicolumn{1}{c|}{53}     & 41           & 37           & 82.3       \\
\textcolor{blue}{$\sbullet[2.0]$}$\dagger$ ResNet 134             & A-A-A              & \multicolumn{1}{c|}{48}     & 38           & 35           & 82.0         \\
\textcolor{blue}{$\sbullet[2.0]$}$\dagger$ ResNet 86              & A-A-A              & \multicolumn{1}{c|}{38}     & 30           & 28           & 81.4       \\
\textcolor{blue}{$\sbullet[2.0]$}$\dagger$ ResNet 56              & A-A-A              & \multicolumn{1}{c|}{31}     & 25           & 22           & 80.7       \\
\textcolor{blue}{$\sbullet[2.0]$}$\dagger$ ResNet 34              & A-A-A              & \multicolumn{1}{c|}{26}     & 21           & 18           & 79.1       \\
\textcolor{blue}{$\sbullet[2.0]$}$\dagger$ ResNet 18              & A-A-A              & \multicolumn{1}{c|}{22}     & 19           & 14           & 76.5       \\ \hline
\textcolor{violet}{$\sbullet[2.0]$}$\dagger$ ResNet 101             & A-A-A              & \multicolumn{1}{c|}{52}     & 36           & 29           & 80.4       \\
\textcolor{violet}{$\sbullet[2.0]$}$\dagger$ ResNet 50              & A-A-A              & \multicolumn{1}{c|}{39}     & 27           & 20           & 79.5       \\ \hline
\textcolor{cyan}{$\sbullet[2.0]$}$\dagger$ ResNet 150             & A-A-C              & \multicolumn{1}{c|}{47}     & 37           & 33           & 81.9       \\
\textcolor{cyan}{$\sbullet[2.0]$}$\dagger$ ResNet 86              & A-A-C              & \multicolumn{1}{c|}{32}     & 26           & 26           & 81.1       \\
\textcolor{cyan}{$\sbullet[2.0]$}$\dagger$ ResNet 34              & A-A-C              & \multicolumn{1}{c|}{21}     & 17           & 16           & 79.1       \\
\textcolor{cyan}{$\sbullet[2.0]$}$\dagger$ ResNet 18              & A-A-C              & \multicolumn{1}{c|}{17}     & 15           & 12           & 76.4       \\ \hline
\textcolor{BlueViolet}{$\sbullet[2.0]$}$\dagger$ ResNet 150 S        & B-B-B              & \multicolumn{1}{c|}{37}     & 31           & 34           &  81          \\
\textcolor{BlueViolet}{$\sbullet[2.0]$}$\dagger$ ResNet 86 S         & B-B-B              & \multicolumn{1}{c|}{26}     & 23           & 26           &  81.1          \\ \hline
\textcolor{Green}{~~~}$\dagger$ ResNet 34              & A-B-B              & \multicolumn{1}{c|}{19}     & 16           & 16           & 78.8       \\
\textcolor{Green}{~~~}$\dagger$ ResNet 18              & A-B-B              & \multicolumn{1}{c|}{15}     & 14           & 12           & 76.7      
\end{tabular}
\end{table}



\textbf{Comparison with DDRNet}:
Other multi-branch architectures such as DDRNet~\cite{ddrnet21} and BiseNet~\cite{bisenet,bisenetv2} have followed HRNet, claiming that it is important to maintain higher resolution information throughout the network, as well inter-branch information exchange. They propose to make such a design faster through reducing the width of the backbone, and maintaining only one higher resolution feature scale instead of multiple. They too, however, use a backbone with a large receptive field. We show comparisons with DDRNet, which has a better reported performance than BiseNets. 

FFNet A-A-A (blue) models using backbones comprised of basic-blocks also perform on par with DDRNets (red), as shown in Figure~\ref{fig:stride_64_models}. As with the case of HRNets, this demonstrates that with a suitably large receptive field, simple encoder-decoder models can work as well as 2-stream models like DDRNets. By changing the backbone and head width of FFNets (blue, black), models that are better than DDRNets can be found for all speed-accuracy tradeoffs. See Table~\ref{tbl:gpu_small}. The 3-stage FFNet-GPU-Large models (green) also outperform DDRNets.

\textbf{Comparison with FANet}:
Approaches like FANet~\cite{fanet2020} and CCNet\cite{ccnet2020} that make use of attention, could be viewed as attempts to increase the effective receptive field of the network, when using backbones with small effective receptive fields. 
Indeed, FANets with bottleneck-blocks in the backbone are comparable to FANets with basic-blocks in the backbone. See Figure~\ref{fig:stride_64_models}.

One of the questions that it naturally leads to is whether an equivalent compute cost increase through enlarging the receptive field of backbone delivers the same accuracy boost that incorporating these attention-based structures does. 
As shown in Figure~\ref{fig:stride_64_models} and Table~\ref{tbl:gpu_small}, FFNet A-A-A (blue) models with a backbone with an adequately large receptive are on-par with FANet (orange, brown) models, and the self-attention head of FANet provides diminishing returns.



\begin{table}[]
\caption{Inference time vs. Cityscapes validaiton accuracy for \textbf{FFNet-Mobile models}: The validation accuracy is for FP16 models, while the inference time is reported for 8 bit quantized models on the Samsung S21 DSP. The inference time is measured for a batch size of 1, with Batch Norm folding. $\dagger$ These models use stride=2 in the first block of the ResNet backbone.}
\label{tbl:device}
\begin{tabular}{cccc}
\multicolumn{1}{c|}{\multirow{3}{*}{Backbone}} & \multicolumn{1}{c|}{\multirow{3}{*}{Stem-Up-Seg}} & \multicolumn{1}{c|}{Inference} & Cityscapes \\
\multicolumn{1}{c|}{}                          & \multicolumn{1}{c|}{}                             & \multicolumn{1}{c|}{Time}      & Val        \\
\multicolumn{1}{c|}{}                          & \multicolumn{1}{c|}{}                             & \multicolumn{1}{c|}{(ms)}      & mIoU       \\ \hline
\multicolumn{4}{c}{Input $1024\times2048$, Output $128\times256$}                                                                                                                    \\ \hline
\multicolumn{1}{l|}{\textcolor{BlueViolet}{$\sbullet[2.0]$}$\dagger$ResNet 78 S}               & \multicolumn{1}{c|}{B-B-B}                        & \multicolumn{1}{c|}{34.0}        & 80.0      \\
\multicolumn{1}{l|}{\textcolor{BlueViolet}{$\sbullet[2.0]$}$\dagger$Resnet 54 S}            & \multicolumn{1}{c|}{B-B-B}                        & \multicolumn{1}{c|}{30.0}        & 79.2       \\
\multicolumn{1}{l|}{\textcolor{BlueViolet}{$\sbullet[2.0]$}$\dagger$ResNet 40 S}            & \multicolumn{1}{c|}{B-B-B}                        & \multicolumn{1}{c|}{27.5}      & 78.8       \\
\multicolumn{1}{l|}{\textcolor{BlueViolet}{$\sbullet[2.0]$}$\dagger$ResNet 30 S}            & \multicolumn{1}{c|}{B-B-B}                        & \multicolumn{1}{c|}{24.0}        & 77.1       \\
\multicolumn{1}{l|}{\textcolor{BlueViolet}{$\sbullet[2.0]$}$\dagger$ResNet 22 S}            & \multicolumn{1}{c|}{B-B-B}                        & \multicolumn{1}{c|}{22.5}      & 75.4       \\ \hline
\multicolumn{4}{c}{Input $512\times1024$, Output $128\times256$}                                                                                                                     \\ \hline
\multicolumn{1}{l|}{\textcolor{blue}{$\sbullet[2.0]$}ResNet 78 S}            & \multicolumn{1}{c|}{B-B-B}                        & \multicolumn{1}{c|}{30.0}          &   79.5         \\
\multicolumn{1}{l|}{\textcolor{blue}{$\sbullet[2.0]$}Resnet 54 S}            & \multicolumn{1}{c|}{B-B-B}                        & \multicolumn{1}{c|}{25.0}          &    79.1        \\
\multicolumn{1}{l|}{\textcolor{blue}{$\sbullet[2.0]$}ResNet 40 S}            & \multicolumn{1}{c|}{B-B-B}                        & \multicolumn{1}{c|}{22.5}      & 78.2       \\
\multicolumn{1}{l|}{\textcolor{blue}{$\sbullet[2.0]$}ResNet 30 S}            & \multicolumn{1}{c|}{B-B-B}                        & \multicolumn{1}{c|}{20.5}      & 76.8       \\
\multicolumn{1}{l|}{\textcolor{blue}{$\sbullet[2.0]$}ResNet 22 S}            & \multicolumn{1}{c|}{B-B-B}                        & \multicolumn{1}{c|}{18.5}      & 74.7       \\ \hline
\multicolumn{1}{l|}{\textcolor{cyan}{$\sbullet[2.0]$}ResNet 78 S}            & \multicolumn{1}{c|}{B-C-C}                        & \multicolumn{1}{c|}{25.0}          & 79.4           \\
\multicolumn{1}{l|}{\textcolor{cyan}{$\sbullet[2.0]$}Resnet 54 S}            & \multicolumn{1}{c|}{B-C-C}                        & \multicolumn{1}{c|}{20.0}          & 78.1           \\
\multicolumn{1}{l|}{\textcolor{cyan}{$\sbullet[2.0]$}ResNet 40 S}            & \multicolumn{1}{c|}{B-C-C}                        & \multicolumn{1}{c|}{18.0}      & 77.5       \\ \hline
\multicolumn{1}{l|}{\textcolor{Green}{$\blacksquare$}ResNet 122 NS}              & \multicolumn{1}{c|}{C-B-B}                        & \multicolumn{1}{c|}{22.0}          & 78.8       \\
\multicolumn{1}{l|}{\textcolor{Green}{$\blacksquare$}ResNet 74 NS}              & \multicolumn{1}{c|}{C-B-B}                        & \multicolumn{1}{c|}{19.0}          & 78.1       \\
\multicolumn{1}{l|}{\textcolor{Green}{$\blacksquare$}ResNet 46 NS}              & \multicolumn{1}{c|}{C-B-B}                        & \multicolumn{1}{c|}{17.0}        & 77.5       \\ \hline
\multicolumn{1}{l|}{\textcolor{YellowGreen}{$\blacksquare$}ResNet 122 NS}              & \multicolumn{1}{c|}{C-C-C}                        & \multicolumn{1}{c|}{18.0}          & 78.7       \\
\multicolumn{1}{l|}{\textcolor{YellowGreen}{$\blacksquare$}ResNet 74 NS}              & \multicolumn{1}{c|}{C-C-C}                        & \multicolumn{1}{c|}{14.0}          & 77.4       \\
\multicolumn{1}{l|}{\textcolor{YellowGreen}{$\blacksquare$}ResNet 46 NS}              & \multicolumn{1}{c|}{C-C-C}                        & \multicolumn{1}{c|}{12.0}      & 76.5      
\end{tabular}
\end{table}




\section{Related Work}
\label{sec:related_work}
For a comprehensive survey of the developments in Deep Learning based Semantic Segmentation, see Minaee et al.~\cite{segmentation_survey}. Here we only cover aspects directly relevant to our discussion.
The go-to architectures for image segmentation roughly belong to two categories. Encoder-decoder designs like UNet~\cite{unet2015}, DeepLabv1/v2/v3~\cite{deeplab17,deeplab2018}, SegNet~\cite{segnet2017}, with a downsampling backbone (encoder) and an upsampling head (decoder), and HRNet-like~\cite{hrnet,liteHRNet21,ddrnet21}, designs which maintain features at multiple resolutions throughout the network.
HRNet-like multi resolution networks typically dominate the results tables, with claims of higher accuracy.

Encoder-decoder designs can have similar sized backbone and head, such as for UNet, or make use of bigger ResNet-like backbones, with various smaller head designs that enable context aggregation, such as for DeepLabv3. 
DeepLabv1/v2/v3 utilize dilated convolutions in the ResNet-like backbones to increase the receptive field size. Dilated convolutions are not efficient on existing hardware. Recent designs such as MobileNetv3~\cite{mobilenetv32019} reduce the use of dilated convolutions by restricting them only to the last block, but do not entirely eliminate them.
Kirillov et al. \cite{panopticFPN} show that the FPN~\cite{FPN2017} design with an un-modified ResNet~\cite{resnet2016} backbone is competitive with DeepLab~\cite{deeplab17} like designs which remove striding and employ dilated convolutions in the ResNet backbone. Panoptic-FPN\cite{panopticFPN} however still employs the ResNet backbones using bottleneck-blocks, a design influenced by the use of smaller images for image classification. 
Our simple encoder-decoder network architecture, FFNet, uses ResNet backbones comprised of basic-blocks, and does not use dilated convolutions.
%
There are various decoder head designs explored in literature, such as Pyramid Pooling, ASPP~\cite{deeplab2018}, with the goal to capture features of different scales. ASPP achieves this through convolutions with different dilations. 
We use a very simple decoder head, and future investigations can determine if substituting it with other head designs is helpful. 
%
%

%
As discussed in Section~\ref{sec:prior_art}, the simple FFNet architecture performs on par with HRNet-like architectures like HRNet~\cite{hrnet}, DDRNet~\cite{ddrnet21}, and GPU-efficient encoder-decoder architectures like FANet~\cite{fanet2020}.

\section{Discussion}
We have only taken a limited look at the avenues of efficiency improvements that can be explored to make FFNets even faster. 
Various approaches to efficient and real-time image segmentation~\cite{bisenet,bisenetv2,mehta2018espnet,liteHRNet21,ddrnet21} make use of FLOP efficient operations and narrower layers, but still within the context of complex meta-architectures. See the related work sections of Hong et al.~\cite{ddrnet21} and Yu et al.~\cite{liteHRNet21} for a detailed discussion of efficient segmentation architectures. These approaches could also be applied to FFNets.

Ideally, the space of efficient architectural configurations should be explored using an automated approach, such as DONNA~\cite{donna2021} or LANA~\cite{LANA2021}. Here too, the simplicity of the design of the meta-architecture is a significant advantage, because the architecture can be trivially split into a small number of simple blocks that can be handled by NAS approaches that rely on blockwise knowledge-distillation~\cite{ofa2020,donna2021,LANA2021}. HRNet, DDRNet etc. either split into a small number of complex blocks, or a very large number of simple blocks, and handling these with blockwise NAS approaches becomes untenable.

Although our study only looks at road-scene semantic segmentation, the simple FFNet architectures could very well be competitive baselines for other computer vision tasks, as well as other semantic segmentation datasets.



\section{Conclusion}

We show that simple FPN-inspired baselines for Semantic Image Segmentation are efficient and extremely competitive with SoTA architectures across a variety of devices, ranging from highly accurate desktop GPU models to highly efficient mobile models. 
We also show that specific architecture instances designed for ImageNet are not necessarily the best for other tasks, and there exist better task specific architectures within the same design space. It is helpful to think about the specific requirements of the task before deciding to port a network from another task.
We hope that this manuscript convinces the reader about the unexplored potential of simpler CNN architectures for semantic segmentation, and beyond.

{\small
\bibliographystyle{ieee_fullname}
\bibliography{segbib}

\begin{thebibliography}{10}\itemsep=-1pt

\bibitem{nvidiasegmentation}
{NVIDIA} semantic segmentation monorepo.
\newblock \url{https://github.com/NVIDIA/semantic-segmentation}.
\newblock Accessed: 2022-03-15.

\bibitem{paperswithcode}
{Papers With Code: State of the Art}.
\newblock \url{https://paperswithcode.com/sota}.
\newblock Accessed: 2022-03-15.

\bibitem{segnet2017}
Vijay Badrinarayanan, Alex Kendall, and Roberto Cipolla.
\newblock Segnet: A deep convolutional encoder-decoder architecture for image
  segmentation.
\newblock {\em IEEE transactions on pattern analysis and machine intelligence},
  39(12):2481--2495, 2017.

\bibitem{ofa2020}
Han Cai, Chuang Gan, Tianzhe Wang, Zhekai Zhang, and Song Han.
\newblock Once for all: Train one network and specialize it for efficient
  deployment.
\newblock In {\em International Conference on Learning Representations}, 2020.

\bibitem{deeplab17}
Liang-Chieh Chen, George Papandreou, Iasonas Kokkinos, Kevin Murphy, and Alan~L
  Yuille.
\newblock Deeplab: Semantic image segmentation with deep convolutional nets,
  atrous convolution, and fully connected crfs.
\newblock {\em IEEE transactions on pattern analysis and machine intelligence},
  40(4):834--848, 2017.

\bibitem{deeplab2018}
Liang-Chieh Chen, Yukun Zhu, George Papandreou, Florian Schroff, and Hartwig
  Adam.
\newblock Encoder-decoder with atrous separable convolution for semantic image
  segmentation.
\newblock In {\em Proceedings of the European conference on computer vision
  (ECCV)}, pages 801--818, 2018.

\bibitem{cordts2016cityscapes}
Marius Cordts, Mohamed Omran, Sebastian Ramos, Timo Rehfeld, Markus Enzweiler,
  Rodrigo Benenson, Uwe Franke, Stefan Roth, and Bernt Schiele.
\newblock The cityscapes dataset for semantic urban scene understanding.
\newblock In {\em Proceedings of the IEEE conference on computer vision and
  pattern recognition}, pages 3213--3223, 2016.

\bibitem{deng2009imagenet}
Jia Deng, Wei Dong, Richard Socher, Li-Jia Li, Kai Li, and Li Fei-Fei.
\newblock Imagenet: A large-scale hierarchical image database.
\newblock In {\em 2009 IEEE conference on computer vision and pattern
  recognition}, pages 248--255. Ieee, 2009.

\bibitem{resnet2016}
Kaiming He, Xiangyu Zhang, Shaoqing Ren, and Jian Sun.
\newblock Deep residual learning for image recognition.
\newblock In {\em Proceedings of the IEEE conference on computer vision and
  pattern recognition}, pages 770--778, 2016.

\bibitem{ddrnet21}
Yuanduo Hong, Huihui Pan, Weichao Sun, and Yisong Jia.
\newblock Deep dual-resolution networks for real-time and accurate semantic
  segmentation of road scenes.
\newblock {\em arXiv preprint arXiv:2101.06085}, 2021.

\bibitem{mobilenetv32019}
Andrew Howard, Mark Sandler, Grace Chu, Liang-Chieh Chen, Bo Chen, Mingxing
  Tan, Weijun Wang, Yukun Zhu, Ruoming Pang, Vijay Vasudevan, et~al.
\newblock Searching for mobilenetv3.
\newblock In {\em Proceedings of the IEEE/CVF International Conference on
  Computer Vision}, pages 1314--1324, 2019.

\bibitem{fanet2020}
Ping Hu, Federico Perazzi, Fabian~Caba Heilbron, Oliver Wang, Zhe Lin, Kate
  Saenko, and Stan Sclaroff.
\newblock Real-time semantic segmentation with fast attention.
\newblock {\em IEEE Robotics and Automation Letters}, 6(1):263--270, 2020.

\bibitem{ccnet2020}
Zilong Huang, Xinggang Wang, Yunchao Wei, Lichao Huang, Humphrey Shi, Wenyu
  Liu, and Thomas~S. Huang.
\newblock Ccnet: Criss-cross attention for semantic segmentation.
\newblock {\em IEEE Transactions on Pattern Analysis and Machine Intelligence},
  2020.

\bibitem{panopticFPN}
Alexander Kirillov, Ross Girshick, Kaiming He, and Piotr Doll{\'a}r.
\newblock Panoptic feature pyramid networks.
\newblock In {\em Proceedings of the IEEE/CVF Conference on Computer Vision and
  Pattern Recognition}, pages 6399--6408, 2019.

\bibitem{FPN2017}
Tsung-Yi Lin, Piotr Doll{\'a}r, Ross Girshick, Kaiming He, Bharath Hariharan,
  and Serge Belongie.
\newblock Feature pyramid networks for object detection.
\newblock In {\em Proceedings of the IEEE conference on computer vision and
  pattern recognition}, pages 2117--2125, 2017.

\bibitem{mehta2018espnet}
Sachin Mehta, Mohammad Rastegari, Anat Caspi, Linda Shapiro, and Hannaneh
  Hajishirzi.
\newblock Espnet: Efficient spatial pyramid of dilated convolutions for
  semantic segmentation.
\newblock In {\em Proceedings of the european conference on computer vision
  (ECCV)}, pages 552--568, 2018.

\bibitem{segmentation_survey}
Shervin Minaee, Yuri~Y. Boykov, Fatih Porikli, Antonio~J Plaza, Nasser
  Kehtarnavaz, and Demetri Terzopoulos.
\newblock Image segmentation using deep learning: A survey.
\newblock {\em IEEE Transactions on Pattern Analysis and Machine Intelligence},
  2021.

\bibitem{LANA2021}
Pavlo Molchanov, Jimmy Hall, Hongxu Yin, Jan Kautz, Nicol{\`{o}} Fusi, and
  Arash Vahdat.
\newblock {LANA:} latency aware network acceleration.
\newblock {\em CoRR}, abs/2107.10624v2, 2021.

\bibitem{donna2021}
Bert Moons, Parham Noorzad, Andrii Skliar, Giovanni Mariani, Dushyant Mehta,
  Chris Lott, and Tijmen Blankevoort.
\newblock Distilling optimal neural networks: Rapid search in diverse spaces.
\newblock In {\em Proceedings of the IEEE/CVF International Conference on
  Computer Vision}, pages 12229--12238, 2021.

\bibitem{unet2015}
Olaf Ronneberger, Philipp Fischer, and Thomas Brox.
\newblock U-net: Convolutional networks for biomedical image segmentation.
\newblock In {\em International Conference on Medical image computing and
  computer-assisted intervention}, pages 234--241. Springer, 2015.

\bibitem{hierarchicalMultiScale2020}
Andrew Tao, Karan Sapra, and Bryan Catanzaro.
\newblock Hierarchical multi-scale attention for semantic segmentation.
\newblock {\em arXiv preprint arXiv:2005.10821}, 2020.

\bibitem{hrnet}
Jingdong Wang, Ke Sun, Tianheng Cheng, Borui Jiang, Chaorui Deng, Yang Zhao,
  Dong Liu, Yadong Mu, Mingkui Tan, Xinggang Wang, Wenyu Liu, and Bin Xiao.
\newblock Deep high-resolution representation learning for visual recognition.
\newblock {\em TPAMI}, 2019.

\bibitem{rw2019timm}
Ross Wightman.
\newblock Pytorch image models.
\newblock \url{https://github.com/rwightman/pytorch-image-models}, 2019.

\bibitem{bisenetv2}
Changqian Yu, Changxin Gao, Jingbo Wang, Gang Yu, Chunhua Shen, and Nong Sang.
\newblock Bisenet v2: Bilateral network with guided aggregation for real-time
  semantic segmentation.
\newblock {\em International Journal of Computer Vision}, 129(11):3051--3068,
  2021.

\bibitem{bisenet}
Changqian Yu, Jingbo Wang, Chao Peng, Changxin Gao, Gang Yu, and Nong Sang.
\newblock Bisenet: Bilateral segmentation network for real-time semantic
  segmentation.
\newblock In {\em Proceedings of the European conference on computer vision
  (ECCV)}, pages 325--341, 2018.

\bibitem{liteHRNet21}
Changqian Yu, Bin Xiao, Changxin Gao, Lu Yuan, Lei Zhang, Nong Sang, and
  Jingdong Wang.
\newblock Lite-hrnet: A lightweight high-resolution network.
\newblock In {\em CVPR}, 2021.

\bibitem{ocrnet19}
Yuhui Yuan, Xilin Chen, and Jingdong Wang.
\newblock Object-contextual representations for semantic segmentation.
\newblock 2020.

\bibitem{rmiLoss2019}
Shuai Zhao, Yang Wang, Zheng Yang, and Deng Cai.
\newblock Region mutual information loss for semantic segmentation.
\newblock {\em Advances in Neural Information Processing Systems}, 32, 2019.

\end{thebibliography}
}
\end{document}